\documentclass[11pt]{article}
\usepackage{blindtext}
\usepackage[preprint,abbrvbib]{jmlr2e}
\usepackage{enumitem}
\usepackage{amsmath}

\usepackage[T1]{fontenc}
\usepackage{listings}
\usepackage{xcolor}
\usepackage{courier}

\definecolor{codegreen}{rgb}{0.15,0.6,0.3}
\definecolor{codegray}{rgb}{0.5,0.5,0.5}
\definecolor{codepurple}{rgb}{0.73,0.13,0.13}
\definecolor{backcolour}{rgb}{1,1,1}
 
\lstdefinestyle{mystyle}{
    backgroundcolor=\color{backcolour},   
    commentstyle=\itshape\color{codegreen},
    keywordstyle=\bfseries\color{magenta},
    numberstyle=\tiny\color{codegray},
    stringstyle=\color{codepurple},
    basicstyle=\footnotesize\ttfamily,
    breakatwhitespace=false,
    breaklines=true,
    captionpos=b,
    keepspaces=true,
    numbers=left,
    numbersep=5pt,
    showspaces=false,
    showstringspaces=false,
    showtabs=false,                  
    tabsize=1
}

\newcommand{\name}{\texttt{ZhiJian}\xspace}

\ShortHeadings{A Modular, Lightweight Toolbox for Evaluating LLM and its Omni-Extensions}{Zhang, Zhong, Lu, Chen, Zhan, and Ye}
\firstpageno{1}

\begin{document}
\lstset{language=Python}

\title{\large{OmniEvalKit: A Modular, Lightweight Toolbox for\\Evaluating Large Language Model and its Omni-Extensions}}

\author{ \\
\name Yi-Kai Zhang$^{1,2}$ \ \ Xu-Xiang Zhong$^{1,2}$ \ \ Shiyin Lu$^3$ \ \ Qing-Guo Chen$^3$ \\ \textbf{De-Chuan Zhan}$^{1,2}$ \ \  \textbf{Han-Jia Ye}$^{1,2}$\thanks{Corresponding author, email: yehj@lamda.nju.edu.cn.} \\
\addr $^1$School of Artificial Intelligence, Nanjing University\\ $^2$National Key Laboratory for Novel Software Technology, Nanjing University\\ $^3$AI Business, Alibaba Group \\}
\editor{}

\maketitle

\begin{abstract}
The rapid advancements in Large Language Models (LLMs) have significantly expanded their applications, ranging from multilingual support to domain-specific tasks and multimodal integration. In this paper, we present \textsc{OmniEvalKit}, a novel benchmarking toolbox designed to evaluate LLMs and their omni-extensions across \textit{multilingual}, \textit{multidomain}, and \textit{multimodal} capabilities. Unlike existing benchmarks that often focus on a single aspect, \textsc{OmniEvalKit} provides a modular, lightweight, and automated evaluation system. It is structured with a modular architecture comprising a Static Builder and Dynamic Data Flow, promoting the seamless integration of new models and datasets.
\textsc{OmniEvalKit} supports over 100 LLMs and 50 evaluation datasets, covering comprehensive evaluations across thousands of model-dataset combinations.
\textsc{OmniEvalKit} is dedicated to creating an ultra-lightweight and fast-deployable evaluation framework, making downstream applications more convenient and versatile for the AI community.
\end{abstract}


\section{Introduction}
The rapid development of Large Language Models (LLMs)~\citep{du2022glm,jiang2023mistral,openai2020chatgpt,touvron2023llama,qwen} has made their question-answering capabilities crucial in many applications. Recently, the inputs for LLMs have been continually expanded to include multiple languages and various specialized domains, including applications worldwide, code generation~\citep{DBLP:journals/corr/abs-2107-03374,DBLP:journals/corr/abs-2308-12950}, mathematical problem-solving~\citep{DBLP:journals/nature/RomeraParedesBNBKDREWFKF24,qwen_2_5_math}, legal inference~\citep{DBLP:conf/emnlp/Fei0ZZHHZ0YSG024}, economic decision-making~\citep{xie2023pixiu}, and medical diagnosis~\citep{DBLP:journals/corr/abs-2304-06975}. Furthermore, Multimodal LLMs~\citep{chen2023sharegpt4v,zhu2023minigpt4,liu2023llava,liu2024llavanext,yao2024minicpm,4v,4o,4oapi} (MLLMs) can integrate diverse forms of information, such as image~\citep{Qwen-VL,Qwen2VL}, video~\citep{zhang2023video-llama,DBLP:conf/slt/ChiCWHC0L21}, or tabular~\citep{DBLP:conf/aistats/HegselmannBLA0S23} inputs.
These advances -- across multilingual, multidomain, and multimodal as M$^3$ \textit{omni-applications} -- are steering us toward Artificial General Intelligence (AGI) systems~\citep{DBLP:journals/jagi/Goertzel14}.

Unlike previous single-task models, LLMs and their extensions are expected to excel in comprehensive zero-shot capabilities, remaining effective in traditional text-only tasks~\citep{DBLP:journals/corr/abs-2406-03496}.
For instance, MLLMs enable real-time translation and enhance cross-cultural communication in international business, while their foundational language also plays a vital role. Domain-specific LLMs can serve as professionals in coding, mathematics, law, finance, or healthcare, requiring a solid foundation in general conversation. Moreover, MLLMs are particularly significant when dealing with interleaved information, especially when processing text-only parts within multimodal content.
Driven by industry demands, LLMs and their omni-extensions have evolved through nearly ten generations, creating hundreds of models across different languages, domains, and modalities. Consequently, fair and comprehensive evaluations are crucial for recognizing model performance in various aspects, making informed subsequent model selections regarding deployment costs and inference overheads. The evaluation process helps identify models with the most robust capabilities and the broadest application prospects.

While many LLM benchmarks are available, most tend to concentrate on a specific type, such as a single language~\citep{singh2024indicqabenchmarkmultilingual,cmmlu}, domain~\citep{DBLP:journals/corr/abs-2309-16289,DBLP:journals/corr/abs-2311-11944}, or modality~\citep{DBLP:conf/mm/DuanYQFCLDZZWL024}. Additionally, the underlying codebases often lack compatibility, and variations in preprocessing configurations and interfaces can be substantial.
For instance, the popular open LLM Leaderboard~\citep{open-llm-leaderboard,open-llm-leaderboard-v2} evaluates text-only question answering but lacks multimodal support due to the fixed default \texttt{generation} method. Similarly, platforms like OpenCompass~\citep{2023opencompass} implement different codes for text-only and multimodal evaluations; however, using multiple codebases for comprehensive evaluation is labor-intensive and may result in biased outcomes.
The phenomenon prompts us to design a modular and lightweight benchmarking toolbox to comprehensively evaluate the diverse and rapidly evolving LLMs and their omni-extensions.

\begin{figure}[t]
    \centering
    \includegraphics[width=0.95\textwidth]{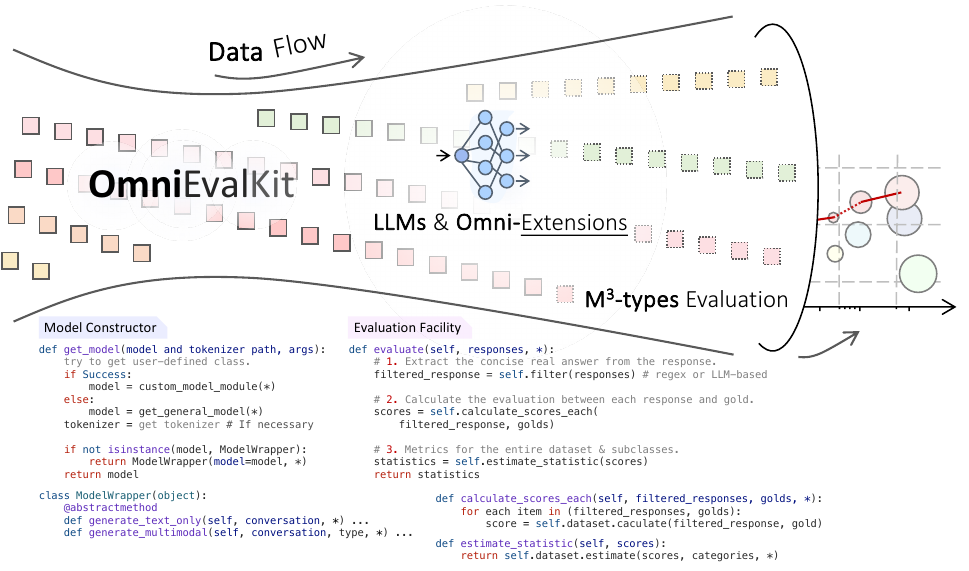}
    \caption{Illustration and API Example of \textsc{OmniEvalKit}.}
    \label{fig:architecture}
\end{figure}

We introduce \textsc{OmniEvalKit}, a unified, comprehensive, and automated framework for evaluating the capability of LLMs and their omni-extensions.
\textsc{OmniEvalKit} expands to include M$^3$ inputs, facilitating a comprehensive evaluation, particularly in text-only \textit{v.s.} visual question-answering tasks, specific languages, and domains of the MLLMs.
The \textsc{OmniEvalKit} framework, as illustrated in~\autoref{fig:architecture}, separates the evaluation process into two primary components: the Static Builder and the Dynamic Data Flow.
The static builder is responsible for constructing the candidate model and establishing evaluation facilities that operate on the dynamic data flow.
For questions in different languages, fields, and modalities, inputs are forwarded into the model, which produces outputs that the evaluation facilities subsequently process. After extracting intent and answers, these outputs are passed to the metrics calculator to obtain results.
The static builder and dynamic data flow decouple the model from the data. The evaluation facilities function like an industrial assembly line, with each component fulfilling a specific role.
This modular architecture facilitates the seamless integration of new components as building blocks in the data flow. It promotes quick extensions and migrations of new models or evaluation datasets via simple interface alignment.
\textsc{OmniEvalKit} supports over 100 LLMs with their omni-extensions and approximately 50 evaluation datasets, enabling evaluations across thousands of model-dataset combinations, in addition to general text-only types with broad natural language tasks, various exams, and specific knowledge items, also includes over 12 general language evaluations, domain-specific evaluations in such as coding, mathematics, law, finance, healthcare, \textit{etc.}, as well as multimodal evaluations with image, video, or tabular inputs.
Due to the highly flexible modularity of \textsc{OmniEvalKit}, it is also compatible with some LLM-related service models, such as vector extraction in retrieval embedding models, model selection, and auxiliary techniques for downstream reuse.
The vision of \textsc{OmniEvalKit} is to build an open evaluation community that offers high availability for users, high accuracy in result efficacy, and comprehensive implementation on deployment.

\section{\textsc{OmniEvalKit} for Comprehensive Evaluation}

In this section, we first introduce the key components of the \textsc{OmniEvalKit} framework. In~\autoref{fig:architecture}, we illustrate how these components are connected through the entire evaluation pipeline using a data flow via the static builder.

\subsection{Key Features}

\begin{itemize}
\item \textbf{Model Types \& Series}: The LLMs and their M$^3$ omni-extensions.
In~\autoref{fig:architecture}, we present a list of models that \textsc{OmniEvalKit} considers, supporting over 100 LLM, as well as LLM's extensions, including the Llama~\citep{touvron2023llama,llama2}, Qwen~\citep{qwen,qwen_2_5,qwen_2_5_math,qwen2_5_code,Qwen-VL,Qwen2VL}, Mistral~\citep{DBLP:journals/corr/abs-2401-04088}, and Phi~\citep{DBLP:journals/corr/abs-2404-14219} series, alongside a variety of proprietary series.
The extended LLMs are proficient at following specific instructions across various language, domain, and modality-based dimensions, while also comprehending general text-only instructions in the original LLM.
\textsc{OmniEvalKit} also accommodates diverse structural inputs, such as image, video, tabular, and other non-text modes.

\item \textbf{Question Types \& Evaluation Benchmarks}: Multiple tasks for general question-answering capabilities, covering multidomain knowledge, multilingual migration, and multimodal information fusion tasks.
We demonstrate that \textsc{OmniEvalKit} is compatible with mainstream evaluation benchmarks, including various examination assessment datasets such as MMLU/CMMLU~\citep{mmlu}, BBH~\citep{DBLP:conf/acl/SuzgunSSGTCCLCZ23}, ARC-Easy/Challenge~\citep{arc}, and OpenbookQA~\citep{DBLP:conf/emnlp/MihaylovCKS18}; typical language datasets like GLUE~\citep{DBLP:conf/iclr/WangSMHLB19}, HellaSwag~\citep{hellaswag}, and ANLI~\citep{DBLP:conf/acl/NieWDBWK20}; and specialized knowledge datasets such as ACLUE~\citep{DBLP:journals/corr/abs-2310-09550}, which focuses on ancient Chinese content, EQ-Bench~\citep{DBLP:journals/corr/abs-2312-06281} for emotional intelligence assessment, and PIQA~\citep{piqa}, which targets commonsense reasoning, among others.
Additionally, \textsc{OmniEvalKit} includes multilingual extensions as MMMLU~\citep{MMMLU}, XStoryCloze~\citep{DBLP:conf/emnlp/LinMAWCSOGBDPSK22}, OALL~\citep{OALL} and multilingual translations of ARC or HellaSwag~\citep{hellaswag}.
Further, \textsc{OmniEvalKit} covers the five major domains of coding, mathematics, law, finance, and healthcare.
It also covers a comprehensive range of general multimodal assessments, including MMMU~\citep{yue2023mmmu}, MME~\citep{fu2023mme}, MMBench~\citep{liu2023mmbench}, MMStar~\citep{DBLP:journals/corr/abs-2403-20330}, and MMVet~\citep{yu2023mmvet}, as well as multimodal-specific scientific question datasets like AI2D~\citep{kembhavi2016ai2d}, ScienceQA~\citep{lu2022scienceqa} and HallusionBench~\citep{guan2023hallusionbench} for capabilities related to specific tasks.

\item \textbf{Answer Extraction Facility}: The omni-extensions of LLMs generate fluent, natural, and human-like responses during deployment, often including additional analysis and connecting words.
For example, some models tend to combine longer thoughts in their responses.
In \textsc{OmniEvalKit}, it supports not only pre-defined rich regular expressions to extract key answers but also the additional LLM to summarize answers from the response.
It features a variety of regularization templates, and the answer extraction model interface aligns with the general one, enabling flexible customization options.

\item \textbf{Model Generation Options}: \textsc{OmniEvalKit} employs a traditional generation approach, incorporating perplexity (PPL) measurements, as MLLM also supports this feature. It provides a flexible assessment interface and offers multiple decoding modes to improve the generation process.

\item \textbf{Accuracy Calculation Center}: 
In addition to the default predefined evaluation metrics such as accuracy, BLEU~\citep{DBLP:conf/acl/PapineniRWZ02}, ROUGE~\citep{lin-2004-rouge}, and others available in the source benchmarks, \textsc{OmniEvalKit} also offers customizable metrics.
It supports various question formats, including single-choice, multiple-choice, yes-or-no, fill-in-the-blank, and free-open ones.

\end{itemize}

\subsection{Evaluation Pipeline}

\begin{itemize}
\item \textbf{Static Builder}: Constructs the LLM or its omni-extension and evaluation facilities.

\begin{itemize}
\item \textbf{Model Constructor}: Configuration component for the evaluated LLM or its extension. The construction and initialization of the model are organized through independent modular files. Customizable interface parameters can be effortlessly passed, enhancing user adaptability. This module is currently responsible for updating the model structure, initializing pretrained parameters, performing GPU mapping, and transforming tokens. Implementing new models is streamlined to focus solely on the relevant model construction class while ensuring alignment with interfaces for token preprocessing, prompt concatenation, and response generation. 
Also, the \texttt{constructor} component supports customization for all generation choices, including different data settings and inference methods. When necessary, default settings are employed to simplify integration and maintain robustness throughout the evaluation.

\item \textbf{Evaluation Facilities}:
Create components dedicated to extracting answers from responses and assessing metrics. This module involves distinct member classes for \texttt{filters} and \texttt{estimators}. The \texttt{filters} remove excessive and redundant information while extracting the critical answers embedded in the model's response. Subsequently, each filtered answer is evaluated against the ground truth by the \texttt{estimators}, and the results are summarized across the entire dataset.

\item \textbf{Other Components}:
Additional parts that assist in the evaluation process. The flexibility of \textsc{OmniEvalKit} allows for the seamless integration of extra processing functions into the overall system. For example, a prompt handling module can be equipped to dynamically concatenate instructions for corresponding additional keys in JSON data files and a token preprocessing unit for effectively reading visual input, such as images, video, or tabular information. These components serve as crucial supportive elements within the overall data flow.
\end{itemize}

\item \textbf{Data Flow}: All datasets are stored in a unified \texttt{JSON} format as a \texttt{list} of \texttt{dicts}. Each dimension, such as domain, language, modality, instruction, and ground truth answer, is recorded in key-value pairs. The file includes different settings for Chain-of-Thought (CoT)~\citep{DBLP:conf/nips/Wei0SBIXCLZ22} and Few-Shot In-Context Learning (FSL, ICL)~\citep{NEURIPS2020_1457c0d6,dong2024surveyincontextlearning}. This standardized and compact format facilitates the expansion of new tasks within the \textsc{OmniEvalKit} framework, enhancing the overall versatility and scalability.
\end{itemize}

\textbf{Data flow-driven interaction with models and evaluation facilities}. When instructions from the data flow are input into the model, the corresponding responses are processed through the answer extraction module and, subsequently, the evaluation module.
Specifically, relevant keys in the data \texttt{JSON} record the prompts required for the instruction, such as in-context few-shot examples or relevant thoughts. As the data flows through the evaluated model, it is concatenated with highly available custom prompts. After inference, the outputs are filtered through a key answer extraction module, where the core content of the responses is extracted using regular expressions or additional models. Subsequently, the estimator module evaluates metrics for each or the entire dataset, aggregating the results obtained.
This systematic approach guarantees coherent and reliable results, preserving both integrity and utility.

\section{Conclusion \& Derivative Fields}

\textsc{OmniEvalKit} is a highly flexible and modular evaluation framework designed for assessing M$^3$ types of LLMs and their omni-extensions. It enables convenient combinations that allow the addition of new models and datasets with just a single-file modification. 
To date, \textsc{OmniEvalKit} has integrated over 50 different evaluation datasets, generating more than 5,000 sets of results for various LLMs and their omni-extensions. This adaptability opens up numerous possibilities for related application areas:
\begin{itemize}
    \item \textbf{New Pattern and Law Exploration}: Investigating the fundamental patterns and laws behind extensive evaluation results.
    For example, scaling laws are crucial for understanding the trends of LLMs and their omni-extensions concerning relevant variables~\citep{DBLP:journals/corr/abs-2001-08361,DBLP:conf/iclr/0006LCF24}, such as model performance relative to the dataset scale and training FLOPs. Scaling laws can guide the selection of hyperparameters and architectures, as well as the prediction of model capabilities for downstream tasks. Leveraging the extensive evaluation results from \textsc{OmniEvalKit}, custom metrics, and training-related performance can be utilized for scaling law research~\citep{DBLP:journals/corr/abs-2001-08361}. In particular, the set of evaluation models quickly deployed within \textsc{OmniEvalKit} and the flexible, adjustable metric assessment module support researchers in exploring the relationships of model capabilities across different dimensions and hierarchical layers for extremely large-scale and diverse model types.

    \item \textbf{Evaluation and Selection of Special Models, Metrics, or Vertical Domain}: When LLMs and their omni-extensions face deployment demands in vertical domains that involve new modalities, domains, and tasks, how to rapidly evaluate existing solutions becomes crucial for guiding pre-selection of optimal models. Some methods~\citep{DBLP:conf/icml/YouLWL21} consider relying on proxy metrics for generalization, measuring them against the evaluation results. Additionally, some learnable strategies~\citep{DBLP:conf/nips/ZhangHDZY23} investigate how to learn generalized criteria for mapping model performance from existing data. Moreover, \textsc{OmniEvalKit} can also be extended to specific model evaluations and data filtering. For example, guiding re-rankers to act as reward models in reinforcement learning from human feedback~\citep{DBLP:conf/nips/Ouyang0JAWMZASR22}, batch-generating data quality paradigms~\citep{pmlr-v202-biderman23a}, or initiating embedding models~\citep{chen2024bge} for a comprehensive assessment of data components.

    \item \textbf{Exploration of New embeddings and Key Outputs}: \textsc{OmniEvalKit}'s modular components continuously evolve, embedding themselves in both the input and output data streams. This framework effectively captures the dynamics of each stage in the model inference process. Not only do these components log the relevant representations produced at any layer, but they also track information regarding future predictions, enabling researchers to gain deeper insights into the model's behavior. Additionally, \textsc{OmniEvalKit} offers customizable decoding methods for generated sequences, creating a flexible environment for the unified evaluation of various search strategies. This comprehensive approach allows researchers to analyze and compare different generation strategies more effectively.
  \end{itemize}

\textsc{OmniEvalKit} has shown stable performance across diverse devices and will continue to evolve to support additional GPU architectures and deep learning deployment frameworks.

\vskip 0.2in
\bibliography{sample}

\begin{thebibliography}{73}
\providecommand{\natexlab}[1]{#1}
\providecommand{\url}[1]{\texttt{#1}}
\expandafter\ifx\csname urlstyle\endcsname\relax
  \providecommand{\doi}[1]{doi: #1}\else
  \providecommand{\doi}{doi: \begingroup \urlstyle{rm}\Url}\fi

\bibitem[Abdin et~al.(2024)Abdin, Jacobs, Awan, Aneja, Awadallah, et~al.]{DBLP:journals/corr/abs-2404-14219}
M.~I. Abdin, S.~A. Jacobs, A.~A. Awan, J.~Aneja, A.~Awadallah, et~al.
\newblock Phi-3 technical report: {A} highly capable language model locally on your phone.
\newblock \emph{CoRR}, abs/2404.14219, 2024.

\bibitem[Bai et~al.(2023{\natexlab{a}})Bai, Bai, Chu, Cui, Dang, et~al.]{qwen}
J.~Bai, S.~Bai, Y.~Chu, Z.~Cui, K.~Dang, et~al.
\newblock Qwen technical report.
\newblock \emph{CoRR}, abs/2309.16609, 2023{\natexlab{a}}.

\bibitem[Bai et~al.(2023{\natexlab{b}})Bai, Bai, Yang, Wang, Tan, Wang, Lin, Zhou, and Zhou]{Qwen-VL}
J.~Bai, S.~Bai, S.~Yang, S.~Wang, S.~Tan, P.~Wang, J.~Lin, C.~Zhou, and J.~Zhou.
\newblock Qwen-vl: A versatile vision-language model for understanding, localization, text reading, and beyond.
\newblock \emph{arXiv preprint arXiv:2308.12966}, 2023{\natexlab{b}}.

\bibitem[Beeching et~al.(2023)Beeching, Fourrier, Habib, Han, Lambert, Rajani, Sanseviero, Tunstall, and Wolf]{open-llm-leaderboard}
E.~Beeching, C.~Fourrier, N.~Habib, S.~Han, N.~Lambert, N.~Rajani, O.~Sanseviero, L.~Tunstall, and T.~Wolf.
\newblock Open llm leaderboard.
\newblock \url{https://huggingface.co/spaces/open-llm-leaderboard-old/open_llm_leaderboard}, 2023.

\bibitem[Biderman et~al.(2023)Biderman, Schoelkopf, Anthony, Bradley, O'Brien, Hallahan, Khan, Purohit, Prashanth, et~al.]{pmlr-v202-biderman23a}
S.~Biderman, H.~Schoelkopf, Q.~G. Anthony, H.~Bradley, K.~O'Brien, E.~Hallahan, M.~A. Khan, S.~Purohit, U.~S. Prashanth, et~al.
\newblock Pythia: A suite for analyzing large language models across training and scaling.
\newblock In \emph{{ICML}}, volume 202 of \emph{Proceedings of Machine Learning Research}, pages 2397--2430, 23--29 Jul 2023.

\bibitem[Bisk et~al.(2020)Bisk, Zellers, Bras, Gao, and Choi]{piqa}
Y.~Bisk, R.~Zellers, R.~L. Bras, J.~Gao, and Y.~Choi.
\newblock {PIQA:} reasoning about physical commonsense in natural language.
\newblock In \emph{{AAAI}}, pages 7432--7439, 2020.

\bibitem[Brown et~al.(2020)Brown, Mann, Ryder, Subbiah, Kaplan, Dhariwal, Neelakantan, Shyam, Sastry, et~al.]{NEURIPS2020_1457c0d6}
T.~Brown, B.~Mann, N.~Ryder, M.~Subbiah, J.~D. Kaplan, P.~Dhariwal, A.~Neelakantan, P.~Shyam, G.~Sastry, et~al.
\newblock Language models are few-shot learners.
\newblock In \emph{{NeurIPS}}, volume~33, pages 1877--1901. Curran Associates, Inc., 2020.

\bibitem[Chen et~al.(2024{\natexlab{a}})Chen, Xiao, Zhang, Luo, Lian, and Liu]{chen2024bge}
J.~Chen, S.~Xiao, P.~Zhang, K.~Luo, D.~Lian, and Z.~Liu.
\newblock Bge m3-embedding: Multi-lingual, multi-functionality, multi-granularity text embeddings through self-knowledge distillation, 2024{\natexlab{a}}.

\bibitem[Chen et~al.(2023)Chen, Li, Dong, Zhang, He, Wang, Zhao, and Lin]{chen2023sharegpt4v}
L.~Chen, J.~Li, X.~Dong, P.~Zhang, C.~He, J.~Wang, F.~Zhao, and D.~Lin.
\newblock Sharegpt4v: Improving large multi-modal models with better captions.
\newblock \emph{arXiv preprint arXiv:2311.12793}, 2023.

\bibitem[Chen et~al.(2024{\natexlab{b}})Chen, Li, Dong, Zhang, Zang, Chen, Duan, Wang, Qiao, Lin, and Zhao]{DBLP:journals/corr/abs-2403-20330}
L.~Chen, J.~Li, X.~Dong, P.~Zhang, Y.~Zang, Z.~Chen, H.~Duan, J.~Wang, Y.~Qiao, D.~Lin, and F.~Zhao.
\newblock Are we on the right way for evaluating large vision-language models?
\newblock \emph{CoRR}, abs/2403.20330, 2024{\natexlab{b}}.

\bibitem[Chen et~al.(2021)Chen, Tworek, Jun, Yuan, de~Oliveira~Pinto, et~al.]{DBLP:journals/corr/abs-2107-03374}
M.~Chen, J.~Tworek, H.~Jun, Q.~Yuan, H.~P. de~Oliveira~Pinto, et~al.
\newblock Evaluating large language models trained on code.
\newblock \emph{CoRR}, abs/2107.03374, 2021.

\bibitem[Chi et~al.(2021)Chi, Chung, Wu, Hsieh, Chen, Li, and Lee]{DBLP:conf/slt/ChiCWHC0L21}
P.~Chi, P.~Chung, T.~Wu, C.~Hsieh, Y.~Chen, S.~Li, and H.~Lee.
\newblock Audio albert: {A} lite bert for self-supervised learning of audio representation.
\newblock In \emph{{IEEE} {SLT}}, pages 344--350, 2021.

\bibitem[Clark et~al.(2018)Clark, Cowhey, Etzioni, Khot, Sabharwal, Schoenick, and Tafjord]{arc}
P.~Clark, I.~Cowhey, O.~Etzioni, T.~Khot, A.~Sabharwal, C.~Schoenick, and O.~Tafjord.
\newblock Think you have solved question answering? try arc, the {AI2} reasoning challenge.
\newblock \emph{CoRR}, abs/1803.05457, 2018.

\bibitem[Contributors(2023)]{2023opencompass}
O.~Contributors.
\newblock Opencompass: A universal evaluation platform for foundation models.
\newblock \url{https://github.com/open-compass/opencompass}, 2023.

\bibitem[Dong et~al.(2024)Dong, Li, Dai, Zheng, Ma, Li, Xia, Xu, Wu, Liu, Chang, Sun, Li, and Sui]{dong2024surveyincontextlearning}
Q.~Dong, L.~Li, D.~Dai, C.~Zheng, J.~Ma, R.~Li, H.~Xia, J.~Xu, Z.~Wu, T.~Liu, B.~Chang, X.~Sun, L.~Li, and Z.~Sui.
\newblock A survey on in-context learning, 2024.
\newblock URL \url{https://arxiv.org/abs/2301.00234}.

\bibitem[Du et~al.(2022)Du, Qian, Liu, Ding, Qiu, Yang, and Tang]{du2022glm}
Z.~Du, Y.~Qian, X.~Liu, M.~Ding, J.~Qiu, Z.~Yang, and J.~Tang.
\newblock Glm: General language model pretraining with autoregressive blank infilling.
\newblock In \emph{{ACL}}, pages 320--335, 2022.

\bibitem[Duan et~al.(2024)Duan, Yang, Qiao, Fang, Chen, et~al.]{DBLP:conf/mm/DuanYQFCLDZZWL024}
H.~Duan, J.~Yang, Y.~Qiao, X.~Fang, L.~Chen, et~al.
\newblock Vlmevalkit: An open-source toolkit for evaluating large multi-modality models.
\newblock In \emph{{ACM} {MM}}, pages 11198--11201. {ACM}, 2024.

\bibitem[Elfilali et~al.(2024)Elfilali, Alobeidli, Fourrier, Boussaha, Cojocaru, Habib, and Hacid]{OALL}
A.~Elfilali, H.~Alobeidli, C.~Fourrier, B.~E.~A. Boussaha, R.~Cojocaru, N.~Habib, and H.~Hacid.
\newblock Open arabic llm leaderboard.
\newblock \url{https://huggingface.co/spaces/OALL/Open-Arabic-LLM-Leaderboard}, 2024.

\bibitem[Fei et~al.(2023)Fei, Shen, Zhu, Zhou, Han, Zhang, Chen, Shen, and Ge]{DBLP:journals/corr/abs-2309-16289}
Z.~Fei, X.~Shen, D.~Zhu, F.~Zhou, Z.~Han, S.~Zhang, K.~Chen, Z.~Shen, and J.~Ge.
\newblock Lawbench: Benchmarking legal knowledge of large language models.
\newblock \emph{CoRR}, abs/2309.16289, 2023.

\bibitem[Fei et~al.(2024)Fei, Shen, Zhu, Zhou, Han, et~al.]{DBLP:conf/emnlp/Fei0ZZHHZ0YSG024}
Z.~Fei, X.~Shen, D.~Zhu, F.~Zhou, Z.~Han, et~al.
\newblock Lawbench: Benchmarking legal knowledge of large language models.
\newblock In \emph{{EMNLP}}, pages 7933--7962, 2024.

\bibitem[Fourrier et~al.(2024)Fourrier, Habib, Lozovskaya, Szafer, and Wolf]{open-llm-leaderboard-v2}
C.~Fourrier, N.~Habib, A.~Lozovskaya, K.~Szafer, and T.~Wolf.
\newblock Open llm leaderboard v2.
\newblock \url{https://huggingface.co/spaces/open-llm-leaderboard/open_llm_leaderboard}, 2024.

\bibitem[Fu et~al.(2023)Fu, Chen, Shen, Qin, Zhang, Lin, Qiu, Lin, Yang, Zheng, et~al.]{fu2023mme}
C.~Fu, P.~Chen, Y.~Shen, Y.~Qin, M.~Zhang, X.~Lin, Z.~Qiu, W.~Lin, J.~Yang, X.~Zheng, et~al.
\newblock Mme: A comprehensive evaluation benchmark for multimodal large language models.
\newblock \emph{arXiv preprint arXiv:2306.13394}, 2023.

\bibitem[Goertzel(2014)]{DBLP:journals/jagi/Goertzel14}
B.~Goertzel.
\newblock Artificial general intelligence: Concept, state of the art, and future prospects.
\newblock \emph{J. Artif. Gen. Intell.}, 5\penalty0 (1):\penalty0 1--48, 2014.

\bibitem[Guan et~al.(2023)Guan, Liu, Wu, Xian, Li, Liu, Wang, Chen, Huang, Yacoob, et~al.]{guan2023hallusionbench}
T.~Guan, F.~Liu, X.~Wu, R.~Xian, Z.~Li, X.~Liu, X.~Wang, L.~Chen, F.~Huang, Y.~Yacoob, et~al.
\newblock Hallusionbench: An advanced diagnostic suite for entangled language hallucination \& visual illusion in large vision-language models.
\newblock \emph{arXiv preprint arXiv:2310.14566}, 2023.

\bibitem[Hegselmann et~al.(2023)Hegselmann, Buendia, Lang, Agrawal, Jiang, and Sontag]{DBLP:conf/aistats/HegselmannBLA0S23}
S.~Hegselmann, A.~Buendia, H.~Lang, M.~Agrawal, X.~Jiang, and D.~A. Sontag.
\newblock Tabllm: Few-shot classification of tabular data with large language models.
\newblock In \emph{{AISTATS}}, volume 206 of \emph{Proceedings of Machine Learning Research}, pages 5549--5581, 2023.

\bibitem[Hendrycks et~al.(2021)Hendrycks, Burns, Basart, Zou, Mazeika, Song, and Steinhardt]{mmlu}
D.~Hendrycks, C.~Burns, S.~Basart, A.~Zou, M.~Mazeika, D.~Song, and J.~Steinhardt.
\newblock Measuring massive multitask language understanding.
\newblock In \emph{{ICLR}}, 2021.

\bibitem[Hendrycks et~al.(2024)Hendrycks, Burns, Basart, Zou, Mazeika, Song, and Steinhardt]{MMMLU}
D.~Hendrycks, C.~Burns, S.~Basart, A.~Zou, M.~Mazeika, D.~Song, and J.~Steinhardt.
\newblock Measuring massive multitask language understanding.
\newblock \url{https://huggingface.co/datasets/openai/MMMLU}, 2024.

\bibitem[Hui et~al.(2024)Hui, Yang, Cui, Yang, Liu, Zhang, Liu, Zhang, Yu, Dang, Yang, Men, Huang, Ren, Ren, Zhou, and Lin]{qwen2_5_code}
B.~Hui, J.~Yang, Z.~Cui, J.~Yang, D.~Liu, L.~Zhang, T.~Liu, J.~Zhang, B.~Yu, K.~Dang, A.~Yang, R.~Men, F.~Huang, X.~Ren, X.~Ren, J.~Zhou, and J.~Lin.
\newblock Qwen2.5-coder technical report.
\newblock \emph{CoRR}, abs/2409.12186, 2024.

\bibitem[Islam et~al.(2023)Islam, Kannappan, Kiela, Qian, Scherrer, and Vidgen]{DBLP:journals/corr/abs-2311-11944}
P.~Islam, A.~Kannappan, D.~Kiela, R.~Qian, N.~Scherrer, and B.~Vidgen.
\newblock Financebench: {A} new benchmark for financial question answering.
\newblock \emph{CoRR}, abs/2311.11944, 2023.

\bibitem[Jiang et~al.(2023)Jiang, Sablayrolles, Mensch, Bamford, Chaplot, de~las Casas, Bressand, Lengyel, Lample, Saulnier, Lavaud, Lachaux, Stock, Scao, Lavril, Wang, Lacroix, and Sayed]{jiang2023mistral}
A.~Q. Jiang, A.~Sablayrolles, A.~Mensch, C.~Bamford, D.~S. Chaplot, D.~de~las Casas, F.~Bressand, G.~Lengyel, G.~Lample, L.~Saulnier, L.~R. Lavaud, M.-A. Lachaux, P.~Stock, T.~L. Scao, T.~Lavril, T.~Wang, T.~Lacroix, and W.~E. Sayed.
\newblock Mistral 7b, 2023.

\bibitem[Jiang et~al.(2024)Jiang, Sablayrolles, Roux, Mensch, Savary, et~al.]{DBLP:journals/corr/abs-2401-04088}
A.~Q. Jiang, A.~Sablayrolles, A.~Roux, A.~Mensch, B.~Savary, et~al.
\newblock Mixtral of experts.
\newblock \emph{CoRR}, abs/2401.04088, 2024.

\bibitem[Kaplan et~al.(2020)Kaplan, McCandlish, Henighan, Brown, Chess, Child, Gray, Radford, Wu, and Amodei]{DBLP:journals/corr/abs-2001-08361}
J.~Kaplan, S.~McCandlish, T.~Henighan, T.~B. Brown, B.~Chess, R.~Child, S.~Gray, A.~Radford, J.~Wu, and D.~Amodei.
\newblock Scaling laws for neural language models.
\newblock \emph{CoRR}, abs/2001.08361, 2020.

\bibitem[Kembhavi et~al.(2016)Kembhavi, Salvato, Kolve, Seo, Hajishirzi, and Farhadi]{kembhavi2016ai2d}
A.~Kembhavi, M.~Salvato, E.~Kolve, M.~Seo, H.~Hajishirzi, and A.~Farhadi.
\newblock A diagram is worth a dozen images.
\newblock In \emph{ECCV}, pages 235--251, 2016.

\bibitem[Li et~al.(2023)Li, Zhang, Koto, Yang, Zhao, Gong, Duan, and Baldwin]{cmmlu}
H.~Li, Y.~Zhang, F.~Koto, Y.~Yang, H.~Zhao, Y.~Gong, N.~Duan, and T.~Baldwin.
\newblock {CMMLU}: Measuring massive multitask language understanding in {Chinese}.
\newblock \emph{arXiv preprint arXiv:2306.09212}, 2023.

\bibitem[Lin(2004)]{lin-2004-rouge}
C.-Y. Lin.
\newblock {ROUGE}: A package for automatic evaluation of summaries.
\newblock In \emph{Text Summarization Branches Out}, pages 74--81. Association for Computational Linguistics, 2004.

\bibitem[Lin et~al.(2022)Lin, Mihaylov, Artetxe, Wang, Chen, et~al.]{DBLP:conf/emnlp/LinMAWCSOGBDPSK22}
X.~V. Lin, T.~Mihaylov, M.~Artetxe, T.~Wang, S.~Chen, et~al.
\newblock Few-shot learning with multilingual generative language models.
\newblock In \emph{{EMNLP}}, pages 9019--9052, 2022.

\bibitem[Liu et~al.(2023{\natexlab{a}})Liu, Li, Wu, and Lee]{liu2023llava}
H.~Liu, C.~Li, Q.~Wu, and Y.~J. Lee.
\newblock Visual instruction tuning.
\newblock \emph{NeurIPS}, 36, 2023{\natexlab{a}}.

\bibitem[Liu et~al.(2024)Liu, Li, Li, Li, Zhang, Shen, and Lee]{liu2024llavanext}
H.~Liu, C.~Li, Y.~Li, B.~Li, Y.~Zhang, S.~Shen, and Y.~J. Lee.
\newblock Llava-next: Improved reasoning, ocr, and world knowledge, January 2024.
\newblock URL \url{https://llava-vl.github.io/blog/2024-01-30-llava-next/}.

\bibitem[Liu et~al.(2023{\natexlab{b}})Liu, Duan, Zhang, Li, Zhang, Zhao, Yuan, Wang, He, Liu, et~al.]{liu2023mmbench}
Y.~Liu, H.~Duan, Y.~Zhang, B.~Li, S.~Zhang, W.~Zhao, Y.~Yuan, J.~Wang, C.~He, Z.~Liu, et~al.
\newblock Mmbench: Is your multi-modal model an all-around player?
\newblock \emph{arXiv preprint arXiv:2307.06281}, 2023{\natexlab{b}}.

\bibitem[Lu et~al.(2022)Lu, Mishra, Xia, Qiu, Chang, Zhu, Tafjord, Clark, and Kalyan]{lu2022scienceqa}
P.~Lu, S.~Mishra, T.~Xia, L.~Qiu, K.-W. Chang, S.-C. Zhu, O.~Tafjord, P.~Clark, and A.~Kalyan.
\newblock Learn to explain: Multimodal reasoning via thought chains for science question answering.
\newblock \emph{NeurIPS}, 35:\penalty0 2507--2521, 2022.

\bibitem[Mihaylov et~al.(2018)Mihaylov, Clark, Khot, and Sabharwal]{DBLP:conf/emnlp/MihaylovCKS18}
T.~Mihaylov, P.~Clark, T.~Khot, and A.~Sabharwal.
\newblock Can a suit of armor conduct electricity? {A} new dataset for open book question answering.
\newblock In \emph{{EMNLP}}, pages 2381--2391. Association for Computational Linguistics, 2018.

\bibitem[Nie et~al.(2020)Nie, Williams, Dinan, Bansal, Weston, and Kiela]{DBLP:conf/acl/NieWDBWK20}
Y.~Nie, A.~Williams, E.~Dinan, M.~Bansal, J.~Weston, and D.~Kiela.
\newblock Adversarial {NLI:} {A} new benchmark for natural language understanding.
\newblock In \emph{{ACL}}, pages 4885--4901. Association for Computational Linguistics, 2020.

\bibitem[OpenAI(2022)]{openai2020chatgpt}
OpenAI.
\newblock Chatgpt.
\newblock \url{https://openai.com/blog/chatgpt}, 2022.

\bibitem[OpenAI(2023)]{4v}
OpenAI.
\newblock Gpt-4v(ision) system card.
\newblock \url{https://cdn.openai.com/papers/GPTV_System_Card.pdf}, 2023.
\newblock Accessed: 2024-05-26.

\bibitem[OpenAI(2024{\natexlab{a}})]{4o}
OpenAI.
\newblock Hello gpt-4o.
\newblock \url{https://openai.com/index/hello-gpt-4o/}, 2024{\natexlab{a}}.
\newblock Accessed: 2024-05-26.

\bibitem[OpenAI(2024{\natexlab{b}})]{4oapi}
OpenAI.
\newblock Introducing gpt-4o: our fastest and most affordable flagship model.
\newblock \url{https://platform.openai.com/docs/guides/vision}, 2024{\natexlab{b}}.
\newblock Accessed: 2024-05-26.

\bibitem[Ouyang et~al.(2022)Ouyang, Wu, Jiang, Almeida, Wainwright, et~al.]{DBLP:conf/nips/Ouyang0JAWMZASR22}
L.~Ouyang, J.~Wu, X.~Jiang, D.~Almeida, C.~L. Wainwright, et~al.
\newblock Training language models to follow instructions with human feedback.
\newblock In \emph{{NeurIPS}}, 2022.

\bibitem[Paech(2023)]{DBLP:journals/corr/abs-2312-06281}
S.~J. Paech.
\newblock Eq-bench: An emotional intelligence benchmark for large language models.
\newblock \emph{CoRR}, abs/2312.06281, 2023.

\bibitem[Papineni et~al.(2002)Papineni, Roukos, Ward, and Zhu]{DBLP:conf/acl/PapineniRWZ02}
K.~Papineni, S.~Roukos, T.~Ward, and W.~Zhu.
\newblock Bleu: a method for automatic evaluation of machine translation.
\newblock In \emph{{ACL}}, pages 311--318. Association for Computational Linguistics, 2002.

\bibitem[Romera{-}Paredes et~al.(2024)Romera{-}Paredes, Barekatain, Novikov, Balog, Kumar, Dupont, Ruiz, Ellenberg, Wang, Fawzi, Kohli, and Fawzi]{DBLP:journals/nature/RomeraParedesBNBKDREWFKF24}
B.~Romera{-}Paredes, M.~Barekatain, A.~Novikov, M.~Balog, M.~P. Kumar, E.~Dupont, F.~J.~R. Ruiz, J.~S. Ellenberg, P.~Wang, O.~Fawzi, P.~Kohli, and A.~Fawzi.
\newblock Mathematical discoveries from program search with large language models.
\newblock \emph{Nat.}, 625\penalty0 (7995):\penalty0 468--475, 2024.

\bibitem[Rozi{\`{e}}re et~al.(2023)Rozi{\`{e}}re, Gehring, Gloeckle, Sootla, Gat, et~al.]{DBLP:journals/corr/abs-2308-12950}
B.~Rozi{\`{e}}re, J.~Gehring, F.~Gloeckle, S.~Sootla, I.~Gat, et~al.
\newblock Code llama: Open foundation models for code.
\newblock \emph{CoRR}, abs/2308.12950, 2023.

\bibitem[Singh et~al.(2024)Singh, Murthy, kumar, Sen, and Ramakrishnan]{singh2024indicqabenchmarkmultilingual}
A.~K. Singh, R.~Murthy, V.~kumar, J.~Sen, and G.~Ramakrishnan.
\newblock Indic qa benchmark: A multilingual benchmark to evaluate question answering capability of llms for indic languages, 2024.
\newblock URL \url{https://arxiv.org/abs/2407.13522}.

\bibitem[Suzgun et~al.(2023)Suzgun, Scales, Sch{\"{a}}rli, Gehrmann, Tay, Chung, Chowdhery, Le, Chi, Zhou, and Wei]{DBLP:conf/acl/SuzgunSSGTCCLCZ23}
M.~Suzgun, N.~Scales, N.~Sch{\"{a}}rli, S.~Gehrmann, Y.~Tay, H.~W. Chung, A.~Chowdhery, Q.~V. Le, E.~H. Chi, D.~Zhou, and J.~Wei.
\newblock Challenging big-bench tasks and whether chain-of-thought can solve them.
\newblock In \emph{{ACL}}, pages 13003--13051. Association for Computational Linguistics, 2023.

\bibitem[Touvron et~al.(2023{\natexlab{a}})Touvron, Lavril, Izacard, Martinet, Lachaux, Lacroix, Rozi{\`e}re, Goyal, Hambro, Azhar, et~al.]{touvron2023llama}
H.~Touvron, T.~Lavril, G.~Izacard, X.~Martinet, M.-A. Lachaux, T.~Lacroix, B.~Rozi{\`e}re, N.~Goyal, E.~Hambro, F.~Azhar, et~al.
\newblock Llama: Open and efficient foundation language models.
\newblock \emph{arXiv preprint arXiv:2302.13971}, 2023{\natexlab{a}}.

\bibitem[Touvron et~al.(2023{\natexlab{b}})Touvron, Martin, Stone, Albert, Almahairi, Babaei, Bashlykov, Batra, Bhargava, et~al.]{llama2}
H.~Touvron, L.~Martin, K.~Stone, P.~Albert, A.~Almahairi, Y.~Babaei, N.~Bashlykov, S.~Batra, P.~Bhargava, et~al.
\newblock Llama 2: Open foundation and fine-tuned chat models.
\newblock \emph{CoRR}, abs/2307.09288, 2023{\natexlab{b}}.
\newblock \doi{10.48550/ARXIV.2307.09288}.
\newblock URL \url{https://doi.org/10.48550/arXiv.2307.09288}.

\bibitem[Wang et~al.(2019)Wang, Singh, Michael, Hill, Levy, and Bowman]{DBLP:conf/iclr/WangSMHLB19}
A.~Wang, A.~Singh, J.~Michael, F.~Hill, O.~Levy, and S.~R. Bowman.
\newblock {GLUE:} {A} multi-task benchmark and analysis platform for natural language understanding.
\newblock In \emph{{ICLR}}. OpenReview.net, 2019.

\bibitem[Wang et~al.(2023)Wang, Liu, Xi, Qiang, Zhao, Qin, and Liu]{DBLP:journals/corr/abs-2304-06975}
H.~Wang, C.~Liu, N.~Xi, Z.~Qiang, S.~Zhao, B.~Qin, and T.~Liu.
\newblock Huatuo: Tuning llama model with chinese medical knowledge.
\newblock \emph{CoRR}, abs/2304.06975, 2023.

\bibitem[Wang et~al.(2024)Wang, Bai, Tan, Wang, Fan, Bai, Chen, Liu, Wang, Ge, Fan, Dang, Du, Ren, Men, Liu, Zhou, Zhou, and Lin]{Qwen2VL}
P.~Wang, S.~Bai, S.~Tan, S.~Wang, Z.~Fan, J.~Bai, K.~Chen, X.~Liu, J.~Wang, W.~Ge, Y.~Fan, K.~Dang, M.~Du, X.~Ren, R.~Men, D.~Liu, C.~Zhou, J.~Zhou, and J.~Lin.
\newblock Qwen2-vl: Enhancing vision-language model's perception of the world at any resolution.
\newblock \emph{arXiv preprint arXiv:2409.12191}, 2024.

\bibitem[Wei et~al.(2022)Wei, Wang, Schuurmans, Bosma, Ichter, Xia, Chi, Le, and Zhou]{DBLP:conf/nips/Wei0SBIXCLZ22}
J.~Wei, X.~Wang, D.~Schuurmans, M.~Bosma, B.~Ichter, F.~Xia, E.~H. Chi, Q.~V. Le, and D.~Zhou.
\newblock Chain-of-thought prompting elicits reasoning in large language models.
\newblock In \emph{{NeurIPS}}, 2022.

\bibitem[Xie et~al.(2023)Xie, Han, Zhang, Lai, Peng, Lopez-Lira, and Huang]{xie2023pixiu}
Q.~Xie, W.~Han, X.~Zhang, Y.~Lai, M.~Peng, A.~Lopez-Lira, and J.~Huang.
\newblock Pixiu: A large language model, instruction data and evaluation benchmark for finance, 2023.

\bibitem[Yang et~al.(2024{\natexlab{a}})Yang, Yang, Hui, Zheng, Yu, et~al.]{qwen_2_5}
A.~Yang, B.~Yang, B.~Hui, B.~Zheng, B.~Yu, et~al.
\newblock Qwen2 technical report.
\newblock \emph{CoRR}, abs/2407.10671, 2024{\natexlab{a}}.

\bibitem[Yang et~al.(2024{\natexlab{b}})Yang, Zhang, Hui, Gao, Yu, Li, Liu, Tu, Zhou, Lin, Lu, Xue, Lin, Liu, Ren, and Zhang]{qwen_2_5_math}
A.~Yang, B.~Zhang, B.~Hui, B.~Gao, B.~Yu, C.~Li, D.~Liu, J.~Tu, J.~Zhou, J.~Lin, K.~Lu, M.~Xue, R.~Lin, T.~Liu, X.~Ren, and Z.~Zhang.
\newblock Qwen2.5-math technical report: Toward mathematical expert model via self-improvement.
\newblock \emph{CoRR}, abs/2409.12122, 2024{\natexlab{b}}.

\bibitem[Yao et~al.(2024)Yao, Yu, Zhang, Wang, Cui, Zhu, Cai, Li, Zhao, He, et~al.]{yao2024minicpm}
Y.~Yao, T.~Yu, A.~Zhang, C.~Wang, J.~Cui, H.~Zhu, T.~Cai, H.~Li, W.~Zhao, Z.~He, et~al.
\newblock Minicpm-v: A gpt-4v level mllm on your phone.
\newblock \emph{arXiv preprint arXiv:2408.01800}, 2024.

\bibitem[You et~al.(2021)You, Liu, Wang, and Long]{DBLP:conf/icml/YouLWL21}
K.~You, Y.~Liu, J.~Wang, and M.~Long.
\newblock Logme: Practical assessment of pre-trained models for transfer learning.
\newblock In \emph{{ICML}}, volume 139, pages 12133--12143, 2021.

\bibitem[Yu et~al.(2023)Yu, Yang, Li, Wang, Lin, Liu, Wang, and Wang]{yu2023mmvet}
W.~Yu, Z.~Yang, L.~Li, J.~Wang, K.~Lin, Z.~Liu, X.~Wang, and L.~Wang.
\newblock Mm-vet: Evaluating large multimodal models for integrated capabilities.
\newblock \emph{arXiv preprint arXiv:2308.02490}, 2023.

\bibitem[Yue et~al.(2023)Yue, Ni, Zhang, Zheng, Liu, Zhang, Stevens, Jiang, Ren, Sun, et~al.]{yue2023mmmu}
X.~Yue, Y.~Ni, K.~Zhang, T.~Zheng, R.~Liu, G.~Zhang, S.~Stevens, D.~Jiang, W.~Ren, Y.~Sun, et~al.
\newblock Mmmu: A massive multi-discipline multimodal understanding and reasoning benchmark for expert agi.
\newblock \emph{arXiv preprint arXiv:2311.16502}, 2023.

\bibitem[Zellers et~al.(2019)Zellers, Holtzman, Bisk, Farhadi, and Choi]{hellaswag}
R.~Zellers, A.~Holtzman, Y.~Bisk, A.~Farhadi, and Y.~Choi.
\newblock {HellaSwag}: Can a machine really finish your sentence?
\newblock In \emph{{ACL}}, pages 4791--4800, 2019.

\bibitem[Zhang et~al.(2024{\natexlab{a}})Zhang, Liu, Cherry, and Firat]{DBLP:conf/iclr/0006LCF24}
B.~Zhang, Z.~Liu, C.~Cherry, and O.~Firat.
\newblock When scaling meets {LLM} finetuning: The effect of data, model and finetuning method.
\newblock In \emph{The Twelfth International Conference on Learning Representations, {ICLR} 2024, Vienna, Austria, May 7-11, 2024}. OpenReview.net, 2024{\natexlab{a}}.

\bibitem[Zhang et~al.(2023{\natexlab{a}})Zhang, Li, and Bing]{zhang2023video-llama}
H.~Zhang, X.~Li, and L.~Bing.
\newblock Video-llama: An instruction-tuned audio-visual language model for video understanding.
\newblock \emph{arXiv preprint arXiv:2306.02858}, 2023{\natexlab{a}}.

\bibitem[Zhang and Li(2023)]{DBLP:journals/corr/abs-2310-09550}
Y.~Zhang and H.~Li.
\newblock Can large language model comprehend ancient chinese? {A} preliminary test on {ACLUE}.
\newblock \emph{CoRR}, abs/2310.09550, 2023.

\bibitem[Zhang et~al.(2023{\natexlab{b}})Zhang, Huang, Ding, Zhan, and Ye]{DBLP:conf/nips/ZhangHDZY23}
Y.~Zhang, T.~Huang, Y.~Ding, D.~Zhan, and H.~Ye.
\newblock Model spider: Learning to rank pre-trained models efficiently.
\newblock In \emph{{NeurIPS}}, 2023{\natexlab{b}}.

\bibitem[Zhang et~al.(2024{\natexlab{b}})Zhang, Lu, Li, Ma, Chen, Xu, Luo, Zhang, Zhan, and Ye]{DBLP:journals/corr/abs-2406-03496}
Y.~Zhang, S.~Lu, Y.~Li, Y.~Ma, Q.~Chen, Z.~Xu, W.~Luo, K.~Zhang, D.~Zhan, and H.~Ye.
\newblock Wings: Learning multimodal llms without text-only forgetting.
\newblock \emph{CoRR}, abs/2406.03496, 2024{\natexlab{b}}.

\bibitem[Zhu et~al.(2024)Zhu, Chen, Shen, Li, and Elhoseiny]{zhu2023minigpt4}
D.~Zhu, J.~Chen, X.~Shen, X.~Li, and M.~Elhoseiny.
\newblock Minigpt-4: Enhancing vision-language understanding with advanced large language models.
\newblock In \emph{{ICLR}}, 2024.

\end{thebibliography}


\begin{thebibliography}{21}
\providecommand{\natexlab}[1]{#1}
\providecommand{\url}[1]{\texttt{#1}}
\expandafter\ifx\csname urlstyle\endcsname\relax
  \providecommand{\doi}[1]{doi: #1}\else
  \providecommand{\doi}{doi: \begingroup \urlstyle{rm}\Url}\fi

\bibitem[Brockman et~al.(2016)Brockman, Cheung, Pettersson, Schneider,
  Schulman, Tang, and Zaremba]{brockman2016openai}
Greg Brockman, Vicki Cheung, Ludwig Pettersson, Jonas Schneider, John Schulman,
  Jie Tang, and Wojciech Zaremba.
\newblock {OpenAI Gym}.
\newblock \emph{arXiv preprint arXiv:1606.01540}, 2016.

\bibitem[Caspi et~al.(2017)Caspi, Leibovich, Novik, and Endrawis]{rlcoach}
Itai Caspi, Gal Leibovich, Gal Novik, and Shadi Endrawis.
\newblock Reinforcement learning {Coach}.
\newblock \url{https://github.com/IntelLabs/coach}, 2017.

\bibitem[Clemente et~al.(2017)Clemente, Castej{\'o}n, and
  Chandra]{clemente2017efficient}
Alfredo~V Clemente, Humberto~N Castej{\'o}n, and Arjun Chandra.
\newblock Efficient parallel methods for deep reinforcement learning.
\newblock \emph{arXiv preprint arXiv:1705.04862}, 2017.

\bibitem[D'Eramo et~al.(2020)D'Eramo, Tateo, Bonarini, Restelli, and
  Peters]{d2020mushroomrl}
Carlo D'Eramo, Davide Tateo, Andrea Bonarini, Marcello Restelli, and Jan
  Peters.
\newblock {MushroomRL}: Simplifying reinforcement learning research.
\newblock \emph{arXiv preprint arXiv:2001.01102}, 2020.

\bibitem[Duan et~al.(2016)Duan, Chen, Houthooft, Schulman, and
  Abbeel]{duan2016benchmarking}
Yan Duan, Xi~Chen, Rein Houthooft, John Schulman, and Pieter Abbeel.
\newblock Benchmarking deep reinforcement learning for continuous control.
\newblock In \emph{International conference on machine learning}, pages
  1329--1338. PMLR, 2016.

\bibitem[Fujita et~al.(2021)Fujita, Nagarajan, Kataoka, and
  Ishikawa]{fujita2021chainerrl}
Yasuhiro Fujita, Prabhat Nagarajan, Toshiki Kataoka, and Takahiro Ishikawa.
\newblock {ChainerRL}: A deep reinforcement learning library.
\newblock \emph{Journal of Machine Learning Research}, 22\penalty0
  (77):\penalty0 1--14, 2021.

\bibitem[Ho and Ermon(2016)]{gail}
Jonathan Ho and Stefano Ermon.
\newblock Generative adversarial imitation learning.
\newblock In \emph{Advances in Neural Information Processing Systems}, pages
  4565--4573, 2016.

\bibitem[Kuhnle et~al.(2017)Kuhnle, Schaarschmidt, and Fricke]{tensorforce}
Alexander Kuhnle, Michael Schaarschmidt, and Kai Fricke.
\newblock Tensorforce: a tensorflow library for applied reinforcement learning.
\newblock \url{https://github.com/tensorforce/tensorforce}, 2017.

\bibitem[Liang et~al.(2018)Liang, Liaw, Nishihara, Moritz, Fox, Goldberg,
  Gonzalez, Jordan, and Stoica]{rllib}
Eric Liang, Richard Liaw, Robert Nishihara, Philipp Moritz, Roy Fox, Ken
  Goldberg, Joseph Gonzalez, Michael~I. Jordan, and Ion Stoica.
\newblock {RLlib}: Abstractions for distributed reinforcement learning.
\newblock In \emph{International Conference on Machine Learning}, pages
  3059--3068, 2018.

\bibitem[Mnih et~al.(2015)Mnih, Kavukcuoglu, Silver, Rusu, Veness, Bellemare,
  Graves, Riedmiller, Fidjeland, Ostrovski, et~al.]{mnih2015human}
Volodymyr Mnih, Koray Kavukcuoglu, David Silver, Andrei~A Rusu, Joel Veness,
  Marc~G Bellemare, Alex Graves, Martin Riedmiller, Andreas~K Fidjeland, Georg
  Ostrovski, et~al.
\newblock Human-level control through deep reinforcement learning.
\newblock \emph{Nature}, 518\penalty0 (7540):\penalty0 529--533, 2015.

\bibitem[Pardo et~al.(2018)Pardo, Tavakoli, Levdik, and
  Kormushev]{pmlr-v80-pardo18a}
Fabio Pardo, Arash Tavakoli, Vitaly Levdik, and Petar Kormushev.
\newblock Time limits in reinforcement learning.
\newblock In \emph{International Conference on Machine Learning}, pages
  4045--4054. PMLR, 2018.

\bibitem[Pathak et~al.(2017)Pathak, Agrawal, Efros, and Darrell]{icm}
Deepak Pathak, Pulkit Agrawal, Alexei~A Efros, and Trevor Darrell.
\newblock Curiosity-driven exploration by self-supervised prediction.
\newblock In \emph{International conference on machine learning}, pages
  2778--2787. PMLR, 2017.

\bibitem[Plappert(2016)]{plappert2016kerasrl}
Matthias Plappert.
\newblock keras-rl.
\newblock \url{https://github.com/keras-rl/keras-rl}, 2016.

\bibitem[Raffin et~al.(2021)Raffin, Hill, Gleave, Kanervisto, Ernestus, and
  Dormann]{stable-baselines3}
Antonin Raffin, Ashley Hill, Adam Gleave, Anssi Kanervisto, Maximilian
  Ernestus, and Noah Dormann.
\newblock {Stable-Baselines3}: Reliable reinforcement learning implementations.
\newblock \emph{Journal of Machine Learning Research}, 2021.

\bibitem[Schulman et~al.(2016)Schulman, Moritz, Levine, Jordan, and
  Abbeel]{gae}
John Schulman, Philipp Moritz, Sergey Levine, Michael~I. Jordan, and Pieter
  Abbeel.
\newblock High-dimensional continuous control using generalized advantage
  estimation.
\newblock In \emph{International Conference on Learning Representations}, 2016.

\bibitem[Silver et~al.(2017)Silver, Schrittwieser, Simonyan, Antonoglou, Huang,
  Guez, Hubert, Baker, Lai, Bolton, et~al.]{alphagozero}
David Silver, Julian Schrittwieser, Karen Simonyan, Ioannis Antonoglou, Aja
  Huang, Arthur Guez, Thomas Hubert, Lucas Baker, Matthew Lai, Adrian Bolton,
  et~al.
\newblock Mastering the game of go without human knowledge.
\newblock \emph{Nature}, 550\penalty0 (7676):\penalty0 354--359, 2017.

\bibitem[Stooke and Abbeel(2019)]{stooke2019rlpyt}
Adam Stooke and Pieter Abbeel.
\newblock rlpyt: A research code base for deep reinforcement learning in
  pytorch.
\newblock \emph{arXiv preprint arXiv:1909.01500}, 2019.

\bibitem[Takuma~Seno(2021)]{seno2021d3rlpy}
Michita~Imai Takuma~Seno.
\newblock d3rlpy: An offline deep reinforcement library.
\newblock In \emph{NeurIPS 2021 Offline Reinforcement Learning Workshop},
  December 2021.

\bibitem[Todorov et~al.(2012)Todorov, Erez, and Tassa]{todorov2012mujoco}
Emanuel Todorov, Tom Erez, and Yuval Tassa.
\newblock {MuJoCo}: A physics engine for model-based control.
\newblock In \emph{International Conference on Intelligent Robots and Systems},
  pages 5026--5033. IEEE, 2012.

\bibitem[van Hasselt et~al.(2016)van Hasselt, Guez, Hessel, Mnih, and
  Silver]{van2016learning}
Hado~P van Hasselt, Arthur Guez, Matteo Hessel, Volodymyr Mnih, and David
  Silver.
\newblock Learning values across many orders of magnitude.
\newblock \emph{Advances in Neural Information Processing Systems},
  29:\penalty0 4287--4295, 2016.

\bibitem[Weng et~al.(2022)Weng, Lin, Huang, Liu, Makoviichuk, Makoviychuk, Liu,
  Song, Luo, Jiang, Xu, and Yan]{envpool}
Jiayi Weng, Min Lin, Shengyi Huang, Bo~Liu, Denys Makoviichuk, Viktor
  Makoviychuk, Zichen Liu, Yufan Song, Ting Luo, Yukun Jiang, Zhongwen Xu, and
  Shuicheng Yan.
\newblock {EnvPool}: A highly parallel reinforcement learning environment
  execution engine.
\newblock \emph{arXiv preprint arXiv:2206.10558}, 2022.

\end{thebibliography}

\end{document}